\documentclass{article}

\usepackage[final,nonatbib]{neurips_2019}

\usepackage{graphicx}
\graphicspath{ {./pics/} }
\usepackage[utf8]{inputenc} 
\usepackage[T1]{fontenc}    
\usepackage{hyperref}       
\usepackage{url}            
\usepackage{booktabs}       
\usepackage{amsfonts}       
\usepackage{nicefrac}       
\usepackage{microtype}      

\usepackage{xcolor}

\title{Focus on What's Informative and Ignore What's not: Communication Strategies in a Referential Game} 

\author{
  Roberto Dess\'{i} \\
  SISSA, Trieste  \\
  Facebook A.I. Research\\
  \texttt{roberto.dessi11@gmail.com} \\
  \And
  Diane Bouchacourt \\
  Facebook A.I. Research \\
  \texttt{dianeb@fb.com} \\
  \AND
  Davide Crepaldi \\
  SISSA, Trieste \\
  \texttt{davide.crepaldi@sissa.it} \\
  \And
  Marco Baroni \\
  Facebook A.I. Research\\
  ICREA \\
  \texttt{mbaroni@fb}
}

\begin{document}

\maketitle

\begin{abstract}
  Research in multi-agent cooperation has shown that artificial
  agents are able to learn to play a simple referential game while
  developing a shared lexicon. This lexicon is not easy to analyze,
  as it does not show many properties of a natural language. In a
  simple referential game with two neural network-based agents, we
  analyze the object-symbol mapping trying to understand what kind of
  strategy was used to develop the emergent language. We see that,
  when the environment is uniformly distributed, the agents rely on a
  random subset of features to describe the objects. When we modify
  the objects making one feature non-uniformly distributed,
  the agents realize it is  less informative and start to ignore it,
  and, surprisingly, they make a better use of the remaining features.
  This interesting result suggests that more natural, less uniformly
  distributed environments might aid in spurring the emergence of
  better-behaved languages.
\end{abstract}

\section{Introduction}



Recent work on language emergence in a multi-agent setup showed that
the interaction between agents performing a cooperative task while
exchanging discrete symbols can lead to the development of a
successful communication system \cite{choi:etal:2018, cao:etal:2018,
  mordatch:abbeel:2017, chaabouni:etal:2019}.  This language protocol
is fundamental to solve the task, since lack of communication results
in performance decrease \cite{lowe:etal:2019}.

One of the simplest configurations is the setup where two artificial
agents play a referential game \cite{lazaridou:etal:2017,
  havrylov:titov:2017, bouchacourt:baroni:2018}.  An agent receives a
target object, and it must communicate about it to another, which is
then tasked to recognize the target in an array of similar items.  The
set of all distinct messages exchanged by the agents is their
communication protocol. The emergent protocol is strictly dependent  
on the distribution of items used in the game.  In fact, when the agents
interact while playing the game, the language gets grounded in the
environment represented by the objects.
Ideally, as it is the case for natural language, we would like to see an
emergent protocol that reflects the structure of the environment.
For instance, when using messages of 3 symbols, if the target object is
a red, full, square and its generated description is
\textit{qry}, \textit{q} could refer to the fact the object is red,
\textit{r} could be related to its texture being full and \textit{y}
could communicate that the object to be described is a square, and this would
suffice to discriminate it among set of similar items.
Similar results have indeed been found in experiments on artificial language evolution
with human subjects \cite{kirby:etal:2008, smith:etal:2003}.

However, analyzing agents language is not trivial, as it does
not necessarily possess properties of natural language such as
consistency or compositionality. Different works have proposed a
qualitative analysis of the object-message mapping performed by the
agents \cite{choi:etal:2018}, others have performed an analysis of the
purity of learned symbols with regards to clustered items 
\cite{lazaridou:etal:2017}.

In this work our goal is to understand if the
agents develop a  communication  strategy that focus on describing salient
features of the objects. We study the symbol-object mapping in a simple
referential game, and how such association changes when the objects in the
environment have different statistical properties, trying to assess how
much information is shared between the object and the emergent protocol.
Borrowing from information theory, mutual information (MI) has been used
to asses form-to-meaning mapping in natural \cite{pimentel:etal:2019} as
well as emergent 
quantifies how many bits of information can we obtain about a distribution
by observing a different one.
Using mutual information between object and message distribution, we show
that when the environment is uniformly distributed, the agents communicate
about an arbitrary subset of the features of the objects. When instead an
environment where some features are less informative than others is used,
the agents learn to ignore those uninformative features, while adapting
to better use a subset of the remaining ones.

\section{Experimental Setup}
Two agents, a sender and a receiver, cooperate in a referential game
where the sender has to describe a discrete-valued target vector to
the receiver, that must distinguish it from a distractor (random
baseline performance is consequently at $50\%$).

\paragraph{Data}
In order to test form-to-meaning mapping variations with objects having 
different distributions, we generate two sets of 5-dimensional
discrete vectors. In one, the feature vectors are uniformly 
distributed, in the other one feature has a highly skewed value distribution.
In the first configuration (uniform environment) each value of the 5
features can be between 1 and 4, each with probability $0.25$. In the
second configuration (skewed environment), the first feature takes
 value $1$ with probability $0.75$, value $2$ with  probability 
$0.15$,  and values $3$ and $4$ with $0.05$ probability.
This leads to configurations with $4^5=1024$ distinct vectors each.
We randomly sampled target-distractor pairs to generate training,
validation and testing partitions (128K, 16K and 4K pairs,
respectively). The test data was used to asses the generalization
capabilities of the agents on unseen pairs.

\paragraph{Game and Network Architectures}
The sender and receiver agents are parametrized with two Vanilla RNNs.
The sender linearly transforms the target vector and feeds it into its
RNN. Through the Gumbel-Softmax trick \cite{maddison:etal:2016,
  jang:etal:2017}, it then generates a message that is passed to the
receiver.  The receiver processes the message through its own RNN
generating an internal representation of the target description. It
linearly processes the pair of vectors it is fed (where target/distractor order is randomized), and it computes a similarity
score through a dot product between each element in the pair and the
message representation.  The highest score is then used to recognize
the target.
Both sender and receiver RNNs are single-layer networks with
50 hidden units and embed the messages into vectors of dimensionality
of 10. For the reparametrization through the Gumbel-softmax, temperature
$\tau$ is kept fixed at 1.

\paragraph{Training and hyperparameter search} We train the agents to
optimize the cross entropy loss using backpropagation and the Adam
optimizer \cite{kingma:ba:2014}.  Vocabulary was fixed to $1100$,
which in our setup ensures that the agent could in principle develop a
one-to-one mapping between input vectors and symbols. The agents were
trained for 50 epochs (after which, they had always converged). We conducted an hyperparameter search varying
batch size (32, 64, 128, 256, 512) and sender and receiver learning
rates (from 0.01, until 0.0001). We cross-validated the top 20
performing models with 5 different initialization seeds, using the
uniform data-set.  The best performing model uses batches of size 64,
with both sender and receiver trained with a learning rate of $0.001$
and an accuracy of $99.04\%$. For both the uniform and skewed
data-sets experiments, we will report results averaged 1,000 runs (10
data-set initialization seeds times 100 network initialization seeeds). A small number of runs did not complete. %
 All experiments were performed using the EGG
toolkit \cite{kharitonov:etal:2019b}. 

\paragraph{Mutual information as a measure of form-to-meaning mapping}
We want to measure to what extent the sender is referring to each feature of the target vector. We measure the consistency in feature-symbol mapping
as the mutual information between each feature distribution and the symbol protocol
distribution as follows \cite{kharitonov:etal:2019a, cover:thomas:2006}:
\begin{equation}
\label{entropy2}
    I(m;v_i) = H(v_i) - H(v_i | m)
\end{equation}
where $H(v_i)$ is the entropy of feature $i$ in the target vectors and $H(v_i | m)$
is the entropy of the conditional distribution of the i-th feature given the messages.
MI is a positive metric, and in our setup it is upper bounded by
the first term of the difference in eq.~\ref{entropy2}, $H(v_i)$. This corresponds 
to a value of $2$ in the uniform data configuration, where each feature in the target
vector can equiprobably take a value between 1 and 4. In the skewed setup, MI is bounded to $2$ for features 2, 3 and 4, and to 
value of $1.15$ for the non-uniform feature.
A high feature-protocol MI implies the agents are consistently using symbols to denote the values of the feature. A low value can
be interpreted as the agents ignoring the feature when describing
the vector.

\section{Results}


\paragraph{Uniform Environment}
Averaged accuracy and MI results are reported in Table~\ref{Table:results}. While showing good performance, the agents do not reach
perfect accuracy, suggesting that they are not making full use of the feature space.
Indeed, although MI has similar average values across features, we constantly
observed that, in each run, three of them had a higher MI compared to the remaining two, in line with our hypothesis that the
agents rely on an arbitrary subset of features to describe the target vector.
In this way, they pay a small accuracy cost in exchange for lower protocol complexity. In the uniform setup, the choice of the feature subset is arbitrary. We looked next at whether we can influence it by making a feature less informative than the others.

\begin{table}[ht!]
    \centering
    \makebox[\textwidth][c]{
        \begin{tabular}{c c c c c c}
            \toprule
            Data & Acc. &  Unique Msgs & H(msgs) & Unique Target Vectors & H(targets) \\
            \cmidrule(r){1-1} \cmidrule(r){2-2} \cmidrule(r){3-3} \cmidrule(r){4-4} \cmidrule(r){5-5} \cmidrule(r){6-6}
            U  &  $98.61\%$  &   50   &   5.46   &  1005 & 9.80 \\
            S  &   $98.52\%$ &   46   &   5.35   &   772 & 8.96 \\
            \midrule
            Data & \multicolumn{5}{c}{Mutual Information}      \\
            \cmidrule(r){1-1} \cmidrule(r){2-6}
            & Feat 1 & Feat 2 & Feat 3 & Feat 4 & Feat 5 \\
            U & 0.62$\pm$0.40 & 0.61$\pm$0.41 & 0.61$\pm$0.41 & 0.60$\pm$0.40 & 0.60$\pm$0.41 \\
            S & 0.07$\pm$0.06 & 0.84$\pm$0.41 & 0.84$\pm$0.42 & 0.82$\pm$0.41 & 0.76$\pm$0.42 \\
            \bottomrule
        \end{tabular}
    }

    \caption{Top: results of the experiments performed with the
      uniform data distribution (U) and with the skewed one
      (S). Unique target vectors and unique msgs represent the distinct
      input vector count seen by the sender and the unique symbols count produced,
      respectively. H(msgs) is the entropy of the emergent protocol,
      the higher the value the more the message distribution approaches
      a uniform one. H(vectors) is the entropy of the target vectors.
      Bottom: mutual information and standard deviation between each
      target vector feature and and the message distribution.
      }    \label{Table:results}
\end{table}

\paragraph{Skewed Environment}
Results are also presented in Table~\ref{Table:results}.
We see that accuracy is comparable in both uniform and non-uniform conditions. The unique target vectors
are less in the skewed configuration than in the uniform one. We also found
a lower entropy in this configuration
as having a non-uniform feature makes sampling the same object more
likely.  The number of unique messages produced, as well as their
entropy, is only very slightly lower in the skewed configuration (8\% less
message for 30\% less target vectors), making us think that messages
are used  effectively. %
Similarly to the uniform environment, also in the the skewed
configuration we see close average values of MI across all the uniform
features, although, again, in specific runs 3 arbitrary features would
have particularly high MIs. We confirm that making one feature in the
environment non-uniform leads the agents to ignore it (MI value of
0.07). More surprisingly, we see higher average MI values for the
remaining features compared to the uniform simulation (features 2, 3
and 4 show a 30\% MI increase, feature 5 a 26\% increase). This
suggests that making one feature less informative has led the agents
to make a better use of the other features, compared to when they have
to choose from a larger set of equally informative ones.


\section{Conclusion and future work}

In order to understand agents' communicative strategy and how it relates to salient
features of the items in the environment,
we presented quantitative evidence of the relation between objects in the environment
and the language protocol developed. Using the referential game setting and tools
from information theory we showed that the agents rely on an arbitrary subset
of the input features to describe target objects.
We also discovered that if we assign a more skewed distribution to a feature of the
environment, they learn to ignore it, as it is less informative in terms of trying
to discriminate a target among similar items. Moreover, with this non-uniform
distribution, the agents adapt to make a better use of the remaining ones. 
As future
work, we plan first of all to understand how consistent this effect is, what causes it,
and under which natural distributions it is more likely to emerge. We want moreover 
to extend the analysis to games using variable length messages, studying how
form-to-meaning mapping might relate to language properties like compositionality.
Another direction will be to analyze the mapping in richer environments having
a greater number of objects and dimensions.
We believe that a better understanding of the object-symbol mapping as well as
of the related strategies used to develop the emergent protocol will be useful to
design better environments to test neural artificial agents and
enforce the emergence of a more human-like language protocol.

\bibliographystyle{abbrv}
\bibliography{roberto}

\end{document}